\documentclass{article}
\usepackage{ijcai26}
\usepackage{times}
\usepackage{soul}
\usepackage{url}
\usepackage[hidelinks]{hyperref}
\usepackage[utf8]{inputenc}
\usepackage[small]{caption}
\usepackage{graphicx}
\usepackage{amsmath}
\usepackage{amsthm}
\usepackage{booktabs}
\usepackage{algorithm}
\usepackage{algorithmic}
\usepackage[switch]{lineno}
\usepackage{amssymb}
\usepackage{multirow}

\usepackage{latexsym}

\title{FunCineForge: A Unified Dataset Toolkit and Model for Zero-Shot Movie Dubbing in Diverse Cinematic Scenes}

\author{
Jiaxuan Liu$^1$\and
Yang Xiang$^1$\and
Han Zhao$^1$\and
Xiangang Li$^1$\and
Zhenhua Ling\\
\affiliations
$^1$Speech Team, Tongyi Lab, Alibaba Group\\
\emails
jxliu@mail.ustc.edu.cn, \{yangxiang.xy, quantu.zh, lixiangang.lxg\}@alibaba-inc.com
}

\begin{document}

\maketitle

\begin{abstract}
    Movie dubbing is the task of synthesizing speech from scripts conditioned on video scenes,
    requiring accurate lip sync, faithful timbre transfer, and proper modeling of character identity and emotion.
    However, existing methods face two major limitations: 
    (1) high-quality multimodal dubbing datasets are limited in scale, suffer from high word error rates, contain sparse annotations, 
    rely on costly manual labeling, and are restricted to monologue scenes,
    all of which hinder effective model training;
    (2) existing dubbing models rely solely on the lip region to learn audio-visual alignment, 
    which limits their applicability to complex live-action cinematic scenes,
    and exhibit suboptimal performance in lip sync, speech quality, and emotional expressiveness.
    To address these issues, we propose FunCineForge, 
    which comprises an end-to-end production pipeline for large-scale dubbing datasets 
    and an MLLM-based dubbing model designed for diverse cinematic scenes.
    Using the pipeline, we construct the first Chinese television dubbing dataset with rich annotations, 
    and demonstrate the high quality of these data. 
    Experiments across monologue, narration, dialogue, and multi-speaker scenes show that 
    our dubbing model consistently outperforms SOTA methods in audio quality, lip sync, timbre transfer, and instruction following.
    Code and demos are available at https://anonymous.4open.science/w/FunCineForge.
\end{abstract}

\section{Introduction}
Movie Dubbing~\cite{V2C}, also known as Visual-Voice-Cloning (V2C), 
is a task that synthesizes speech for video clips from scripts using target timbres. 
This technique has significant potential in television and film production, digital media, and video editing.
Unlike Text-to-Speech (TTS), movie dubbing requires incorporating visual information from character faces to precisely control the duration, emotion, and prosody of the synthesized speech.

Existing movie dubbing methods can be broadly categorized into three research paradigms.
The first category focuses on improving the synchronization between the synthesized speech and lip movements
by learning visual features from the speaker's lip region. 
For example, a text-video aligner is introduced in NeuralDubber~\cite{NeuralDubber}, a duration aligner is introduced in HPMDubber~\cite{HPMDubber},
and a phoneme-guided lip aligner is introduced in StyleDubber~\cite{StyleDubber}.
These methods aim to explicitly control phoneme-level speech duration by mapping textual phonemes to video frames.
However, their alignment performance heavily relies on the presence of clear and visible lip regions,
making them effective only for simple, frontal, and high-resolution monologue scenes.
In live-action television, complex cinematic scenes such as 
facial occlusion, frequent shot changes, multiple coexisting speakers, large speaker-camera distances, and low-resolution frames are common, 
which prevents reliable audio-visual alignment.

\begin{figure}
    \centering
    \centerline{\includegraphics[width=\linewidth]{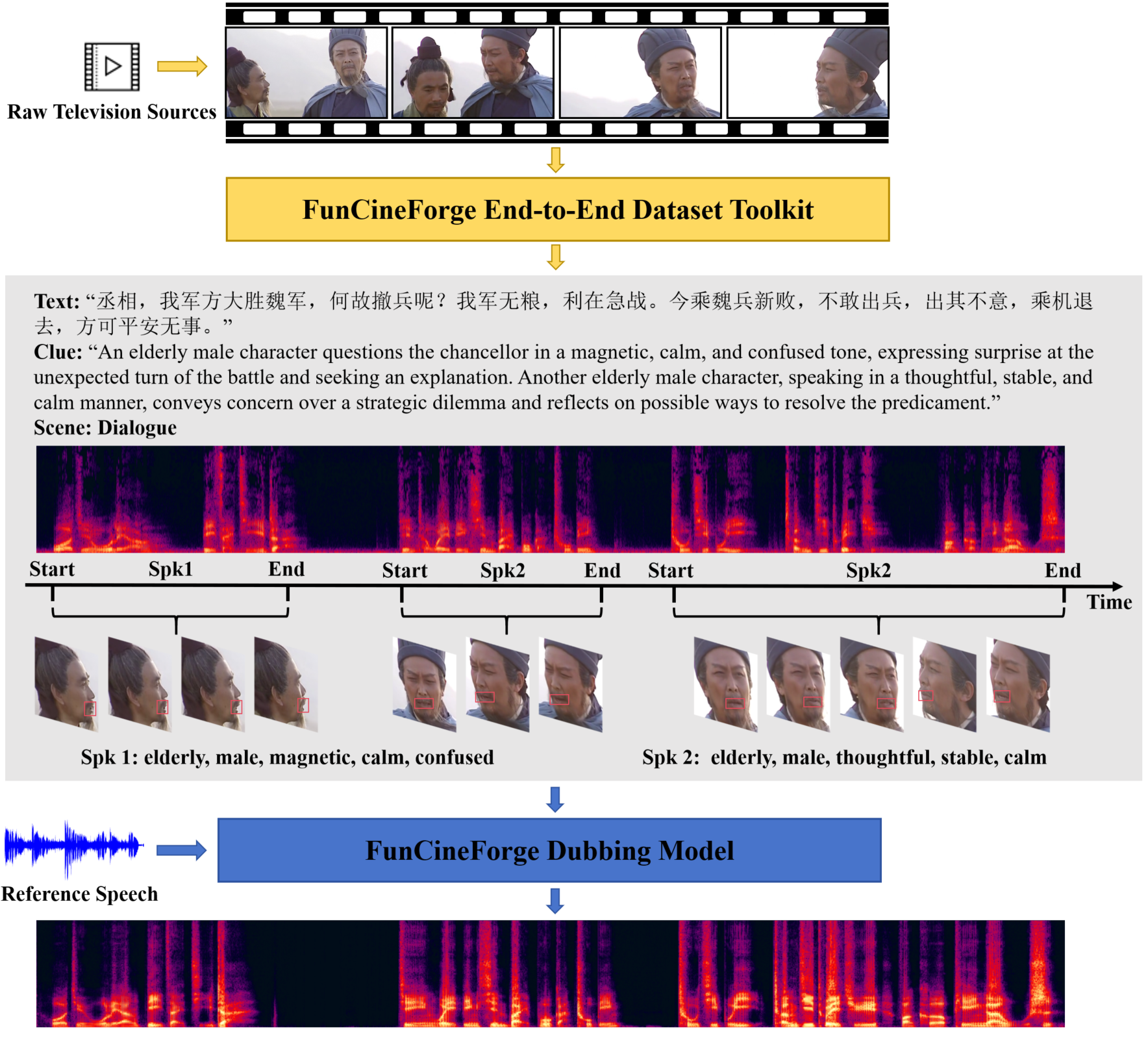}}
    \caption{The overview of FunCineForge. 
    The dataset pipeline automatically transforms raw film and television sources into structured multimodal data, 
    which are used to train and evaluate the dubbing model.
    During inference, given a silent video clip, dubbing script, clue instructions, scene category, and reference speeches,
    the model synthesizes scene-consistent speech.}
    \label{fig:overview}
\end{figure}

The second category focuses on enhancing dubbing quality and speaker similarity.
For example, Speaker2Dubber~\cite{Speaker2Dubber} improves synthesis quality by pre-training the phoneme encoder 
on a large-scale TTS corpus before on dubbing datasets.
FlowDubber~\cite{FlowDubber} introduces dual contrastive learning and LLM-based flow matching guidance, significantly enhancing speech quality.
ProDubber~\cite{ProDubber} proposes prosody-enhanced acoustic pretraining and acoustic-disentangled prosody adaptation
to improve prosody modeling and expressiveness.
EmoDubber~\cite{EmoDubber} employs a flow-based user emotion controlling with positive and negative
guidance mechanisms to manipulate dubbing emotions.
Authentic-Dubber~\cite{AuthenticDubber} enhances emotional retrieval and matching by constructing a multimodal reference footage library and employing emotion-similarity-based retrieval augmentation.
To better clone the reference speaker timbre, most above works adopt acoustic encoders to extract speaker embeddings,
while Lee et al.~\shortcite{ImaginaryVoice} learn speaker identity from facial features to capture speaker paralinguistic information.
However, these methods heavily rely on clear, single-speaker faces and consistent timbre.
In the presence of shot-level facial transitions, speaker timbre switching, or intermediate silent segments, 
alignment becomes difficult, and both dubbing quality and speaker similarity degrade substantially.

The third category leverages Multimodal Large Language Models (MLLMs) to fuse information from multiple modalities for movie dubbing. 
DeepDubber-V1~\cite{DeepDubber-V1} utilizes multimodal Chain-of-Thought (CoT) reasoning on visual inputs 
to infer dubbing styles and speaker attributes, 
generating visual-derived clues that serve as instructions for the dubbing MLLM.
InstructDubber~\cite{InstructDubber} enables controllable dubbing by conditioning the model on textual prompts 
that describe speaker identity and emotional clues as an additional modality.
These methods primarily rely on natural language instructions to control speech style and emotion, 
offering a more intuitive and interpretable paradigm for movie dubbing.
However, the effectiveness of MLLMs critically depends on large-scale dubbing datasets, whereas existing datasets are limited in scale, 
confined to animated content, and lack rich annotations, which significantly constrains their effectiveness in real-world scenarios.

To address these issues, we propose FunCineForge,
which comprises an end-to-end production pipeline for large-scale dubbing datasets 
and an MLLM-based dubbing model designed for diverse cinematic scenes, as illustrated in Fig.~\ref{fig:overview}.
The FunCineForge dataset pipeline combines the advantages of lightweight specialized models and general-purpose MLLMs,
innovatively adopting {\bf specialized models prediction plus Multimodal CoT Correction} strategy.
It automatically transforms raw film and television sources into structured multimodal data, 
including text transcripts, vocal and instrumental tracks, video streams, 
character and emotion clues, dialogue annotations (speaker gender, age, timbre, and start-end timestamps), 
facial and lip sequences, and scene categories.
Using the pipeline, we construct {\bf CineDub-CN}, the first large-scale Chinese television dubbing dataset, with over 4,700~hours of data.
It features accurate text transcripts, reasonable distributions of scene categories and speaker age and gender, 
generalized timbre keywords, robust multi-scale clue annotations, and frame-level accurate timestamps.
Based on this dataset, we propose the FunCineForge dubbing model, 
which integrates more multimodal information within an MLLM architecture.
We propose a novel multimodal alignment mechanism that jointly models {\bf where speech can occur, what is spoken, and ﬁne-grained lip-speech alignment},
together with an improved flow matching module that {\bf supports speaker switching}.
As a result, the proposed dubbing model synthesizes highly natural speech with precise lip sync,
effectively handling complex cinematic scenes and flexible speaker switching.
We release part of our source code, synthesized demos and CineDub-CN dataset at \url{https://anonymous.4open.science/r/FunCineForge}.

\section{Related Works}
\subsection{Dubbing Datasets}
Dubbing datasets used for model training, including V2C-Animation~\cite{V2C}, Chem~\cite{Chem}, and GRID~\cite{GRID}, are limited in scale. 
Even the largest one, V2C-Animation, contains only 10,217 video-audio-text clips with an average duration of 2.4 seconds, totaling 6.8 hours of data.
These datasets heavily rely on manual annotation or crowd-sourced services, are difficult to scale, and are limited to monologue scene due to excessively short clips.
The scarcity of data and limited, inconsistent annotations also hinder the effectiveness of MLLMs in complex cinematic scenes,
highlighting the urgent need for a fully automated pipeline to construct large-scale dubbing datasets.

\subsection{LLM-based Text-to-Speech}
Recent advances in large language models (LLMs) have reshaped the field of TTS.
Most LLM-based TTS methods adopt a two-stage paradigm,
where an LLM first generates speech semantic tokens from text and an acoustic decoder converts them into acoustic features such as Mel-spectrograms,
achieving significant advantages in synthesizing expressive speech.
Representative models include CosyVoice 3~\cite{CosyVoice3}, IndexTTS 2~\cite{IndexTTS2}, 
FireRedTTS 2~\cite{FireRedTTS2}, VibeVoice~\cite{VibeVoice} and Spark-TTS~\cite{Spark-TTS}, 
In parallel, flow matching has emerged as an effective acoustic decoding strategy.
Matcha-TTS~\cite{Matcha-TTS} adopts optimal-transport conditional flow matching for single-speaker TTS, 
while CosyVoice 2~\cite{CosyVoice2} demonstrates the effectiveness of integrating flow matching with LLMs.

Despite their success, these methods cannot provide the speech with desired emotion and audio-visual sync for movie dubbing.
Moreover, existing methods primarily focus on single-speaker timbre cloning
and struggle with speaker switching in dialogue and multi-speaker cinematic scenarios.

\begin{figure*}[!t]
    \centering
    \centerline{\includegraphics[width=\linewidth]{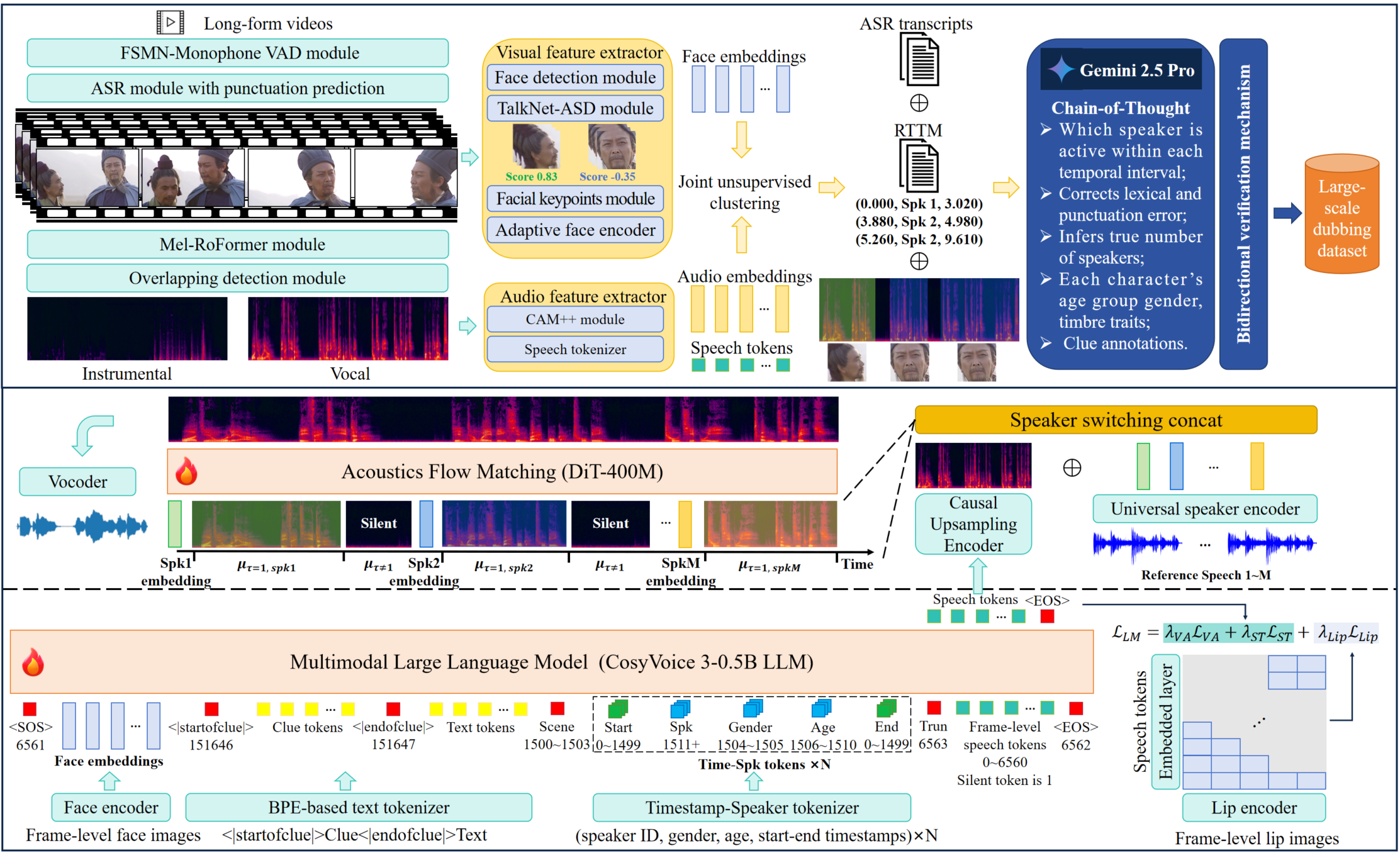}}
    \caption{
        The main architecture of FunCineForge. The framework consists of a dataset pipeline and a dubbing model.
        The dubbing model comprises two components: 
        an MLLM for multimodal alignment, and
        a flow matching module with speaker switching.
    }
    \label{fig:FunCineForge}
\end{figure*}

\subsection{Speaker Diarization for Cinematic Scenes}
Speaker Diarization task~\cite{SD} aims to segment an audio recording into regions corresponding to different speakers
by estimating both the number of speakers and the start-end timestamps of their speech segments.
In complex cinematic scenes with frequent speaker switching or multi-speaker interactions, 
accurate movie dubbing requires explicit knowledge of which speaker is active at each temporal interval, 
making robust speaker diarization a crucial prerequisite.
Most speaker diarization methods~\cite{SD-2,SD-3} follow a feature extraction and unsupervised clustering pipeline,
consisting of: (1) voice activity detection (VAD) to remove non-speech regions; 
(2) sliding-window extraction of speaker embeddings using models such as Res2Net~\cite{Res2Net} and CAM++~\cite{CAM++}; 
and (3) unsupervised clustering of speech segments based on speaker embeddings.
3D-Speaker~\cite{3D-Speaker} further incorporates visual embeddings for visually enhanced clustering,
achieving notable performance gains.

Despite these advances, speaker diarization in long-form cinematic content remains challenging due to:
(1) clustering errors caused by a large number of speakers in long-form videos; 
(2) failures under overlapping multi-speaker speech; and 
(3) speaker confusion induced by inconsistencies between the number of visible speakers and the number of active audio speakers~\cite{SpeakerLM}. 
Consequently, effective segmentation plus speaker diarization for long-form cinematic scenes remain open problems.

\section{FunCineForge}
\subsection{Dataset Pipeline}
To enable training of dubbing models in diverse cinematic scenes, the FunCineForge dataset pipeline transforms raw long-form television sources into structured multimodal data. 
It is specifically designed to address challenges such as high character error rates in transcription and inaccurate speaker diarization.
As shown in Fig.~\ref{fig:FunCineForge}, the pipeline comprises several coordinated workflows, summarized in four stages.

\subsubsection{Video Standardization and Segmentation}
For long-form videos, an FSMN-Monophone VAD~\cite{VAD} module optimized for long-duration speech is first applied to extract speech-active segments. 
These segments are then transcribed using an automatic speech recognition (ASR)~\cite{FunASR} module and 
further split into sentence-level units based on punctuation, 
producing sentence-level SubRip Subtitle (SRT) files.
Based on the obtained timestamps, the original video is subsequently segmented into shorter video-audio clips.

\subsubsection{Vocal Separation and Filtering}
To improve vocal clarity and suppress background music, 
thereby lowering character error rates and facilitating reliable speaker embedding extraction, 
a Mel-RoFormer~\cite{Mel-RoFormer} module is employed to separate vocal and instrumental components. 
This process effectively removes music and background interference from audio mixtures and yields clean vocal tracks. 
Subsequently, overlapped speech detection module is applied to the vocal tracks 
to identify and discard video-audio clips containing overlapping speakers, 
ensuring the reliability of downstream speaker diarization.

\subsubsection{Audio-Visual Speaker Diarization}
The pipeline adopts a visually enhanced speaker diarization framework.
In the audio modality, CAM++~\cite{CAM++} extracts audio speaker embeddings from clean vocal tracks 
and the CosyVoice 3 speech tokenizer~\cite{CosyVoice3} encodes audio into speech tokens at 25 Hz.
In the visual modality, frame sampling is first performed by selecting one frame every five frames for 25 fps videos.
All faces in these sampled frames are detected
using a lightweight face detection module\footnote{https://github.com/Linzaer/Ultra-Light-Fast-Generic-Face-Detector-1MB}
to obtain candidate active frames.
These faces with audio embeddings are then scored by TalkNet-ASD\footnote{https://github.com/TaoRuijie/TalkNet-ASD} module, 
and the face with the highest score in each active frame is selected as the active speaker.
For the selected face, a FAN-based~\cite{FAN} module is applied to detect 2D facial keypoints, 
from which lip coordinates are obtained and raw face and mouth regions are extracted.
An adaptive face encoder~\cite{CurricularFace} is used to extract face embeddings of the active speaker, 
followed by normalization to suppress expression-related variations and retain identity-related representations.
Finally, a joint unsupervised clustering strategy~\cite{3D-Speaker} is applied, 
where audio speaker embeddings serve as the primary modality and 
normalized visual face embeddings act as auxiliary cues, 
producing robust speaker clustering results formatted as Rich Transcription Time Marked (RTTM) files.

\subsubsection{Multimodal CoT Correction with MLLM}
To provide the FunCineForge dubbing model with auxiliary paralinguistic information as instructions,
it is necessary to enrich each video-audio clip with holistic descriptions of character attributes and emotional tone.
Traditionally, such annotations are manually curated at high cost.
Recent advances in general-purpose MLLMs, such as GPT-5, Gemini-2.5, Qwen-3, 
enable their automatic generation.
In addition, ASR transcripts often suffer from word omissions, word errors and punctuation mistakes,
while speaker diarization results may contain duplicated or missing speaker identities.
To further improve dataset quality, the pipeline introduces an MLLM-based correction strategy.

Specifically, the pipeline employs Gemini-2.5-Pro~\cite{Gemini2.5}, 
an MLLM with advanced audio understanding for multimodal CoT reasoning.
The inputs include clean vocal tracks, ASR transcripts, and speaker diarization tuples
in the form of \textit{(start timestamp, speaker ID, end timestamp)}.
The MLLM first understands which speaker is active within each temporal interval and 
corrects lexical and punctuation errors in the ASR transcripts.
Based on its understanding of the audio content, CoT reasoning is then used to infer
the true number of speakers, along with each character's gender, age group, timbre traits, and the overall emotional tone.
Finally, these attributes are summarized into structured long-span and short-span clue annotations,
covering both character profiles and emotional descriptions.

To ensure consistency and reliability, the pipeline enforces strict prompt templates for MLLM generation.
It further integrates lightweight specialized models with general-purpose MLLMs through a bidirectional verification mechanism
to address occasional instability and hallucination in MLLM outputs.
The MLLM outputs are parsed and normalized via front-end processing, such as traditional-to-simplified Chinese conversion and numeric normalization.
Samples with a Levenshtein distance greater than 50\% between the MLLM-corrected transcripts and the ASR transcripts are discarded,
while low-discrepancy samples retain the corrected transcripts.
Similarly, samples with inconsistent speaker identities between MLLM correction and specialized models predictions are removed,
and unreliable gender and age labels are replaced with \textit{Unknown}.
Finally, samples are categorized into monologue, narration, dialogue, and multi-speaker scenes
based on the presence of facial data and the number of active speakers.

\begin{figure}
    \centering
    \centerline{\includegraphics[width=\linewidth]{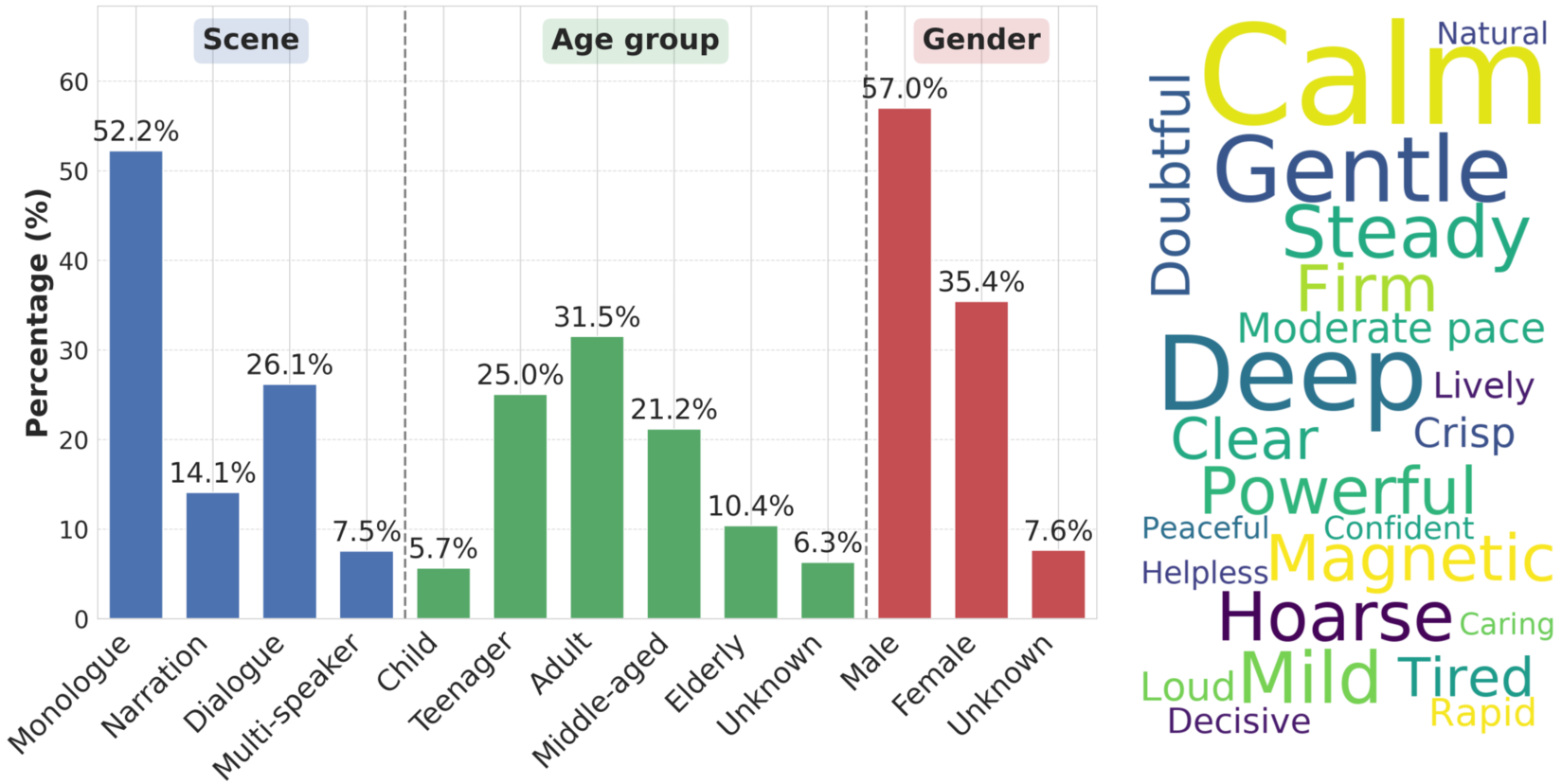}}
    \caption{The left subfigure shows the distributions of scene categories, speaker age groups, and gender, 
    while the right subfigure presents a timbre-related keyword cloud of the CineDub-CN dataset.}
    \label{fig:statistics}
\end{figure}
\subsection{Dubbing Model}
The architecture of dubbing model is illustrated in Fig.~\ref{fig:FunCineForge}.
Formally, given an input video from which facial frame sequences $Face\in\mathbb{R}^T$ of the active speaker are extracted, 
where $T$ denotes the number of frames, a dubbing script $Text$, a clue instruction $Clue$,
a scene category $Scene\in\mathbb{R}^4$ corresponding to monologue, narration, dialogue, and multi-speaker,
a set of timestamp-speaker tuples $\mathcal{T}^N=(start, spk, gender, age, end)^N$ describing $N$ non-silent segments,
and reference speech samples $ref^{M}$ from $M$ speakers,
the model synthesizes the Mel-spectrograms,
\begin{equation}
    \hat{Y}=\text{Model}(Face,Text,Clue,Scene,\mathcal{T}^N,ref^M).
\end{equation}

The dubbing model is trained in two stages: MLLM training and flow matching training.
After flow matching, the synthesized Mel-spectrograms $\hat{Y}$ are converted into waveforms using the widely used HiFiGAN vocoder.

\subsubsection{MLLM with Multimodal Alignment}

First, the dubbing script and clue instruction are concatenated as
"$\langle|$startofclue$|\rangle Clue\langle|$endofclue$|\rangle Text$",
and tokenized using a byte pair encoding (BPE)-based tokenizer to obtain the combined text tokens $X_{Text}$.
Face and lip images in $Face$ are extracted from 25-fps videos and uniformly sampled every five frames, resulting in 5-fps visual sequences.
These downsampled face and lip frames are efficiently encoded into sparse facial and lip representations 
using the face encoder~\cite{CurricularFace} and a lip encoder~\cite{HPMDubber}, respectively, 
without normalization, thereby preserving expression-related variations.
The resulting features are then upsampled to match the original video frame rate, 
producing face embeddings $E_{Face}^T$ and lip embeddings $E_{Lip}^T$
with the same sequence length as the speech tokens $X_{Speech}$.

To handle frequent shot changes, speaker switching, facial occlusion and facial ambiguity in complex cinematic scenes, 
and to achieve accurate audio-visual alignment, 
we design a multimodal alignment mechanism that combines strong and weak supervision
to jointly model where speech can occur, what is spoken, and fine-grained lip-speech alignment.

(1) The provided timestamp-speaker tuples $\mathcal{T}^N$ serve as strong supervision,
explicitly constraining the temporal intervals \emph{where speech can occur} and enabling the MLLM to learn precise temporal alignment.
A dedicated frame-index codebook is introduced to encode start-end timestamps.
Since each video clip is limited to 60 seconds and sampled at 25 fps, the maximum length of timestamp tokens is 1500.
To jointly encode temporal and speaker-related information, 
we design a Timestamp-Speaker tokenizer (TST) that maps timestamp and speaker attributes into a discrete token sequence $X_{TS}$.
Specifically, each segment is represented by tokens corresponding to speaker identity $X_{Spk}$, 
gender token $X_{Gender}$ for male and female, 
and $X_{Age}$ corresponding to child, teenager, adult, middle-aged, and elderly age groups.
Samples with \textit{Unknown} annotations are masked at the attribute-token during training.
In addition, the silent regions with token value of 1, 
we define a frame-level voice activity indicator as
$\tau_t=1$ if $X_{Speech}^{(t)} \neq 1$, otherwise $\tau_t=0$,
where $\tau_t$ denotes the ground-truth voice activity at frame $t$, 
and $\hat{\tau}_t$ represents the predicted voice activity derived from the synthesized speech tokens.
A voice activity loss is constructed over $X_{Speech}$ as
\begin{equation}
    \mathcal{L}_{VA}=-\sum_{t=1}^{T}[\tau_t \log(\hat{\tau_t}) + (1-\tau_t)\log(1-\hat{\tau_t})].
\end{equation}

(2) For speech tokens $X_{Speech}$, the supervision mainly focuses on \emph{what is spoken}.
Accordingly, the speech token loss is defined as a cross-entropy loss under conditions $\mathcal{C}_{LM}$,
\begin{equation}
    \mathcal{L}_{ST}=-\frac{1}{T+1}\sum_{t=1}^{T+1} \log p(X_{Speech}^{(t)} \mid X_{Speech}^{(<t)}, \mathcal{C}_{LM}).
\end{equation}

(3) Visual features are introduced to enhance the MLLM's understanding of speaker identity and emotional expression,
and to provide weak supervision for \emph{fine-grained lip-speech alignment} when lip movements are clearly observable.
Specifically, frame-level face embeddings $E_{Face}^T$, temporally aligned with the speech tokens $X_{Speech}$, are fed into the MLLM.
A contrastive learning objective is introduced between frame-level lip embeddings $E_{Lip}^T$ and 
speech token embeddings $E_{ST}^T=g_{st}(Emb(X_{Speech}))$,
\begin{equation}
    \mathcal{L}_{Lip}=-\sum_{t=1}^T\tau_tw_t\log\frac{
            \exp\left(\langle E_{Lip}^{(t)},E_{ST}^{(t)} \rangle / \tau\right)
        }{
            \sum_{s=1}^{T}\exp\left(\langle E_{Lip}^{(t)},E_{ST}^{(t)}\rangle / \tau\right)
        },
\end{equation}
where $\tau_t$ is activated within non-silent intervals,
and $w_t$ down-weights frames with weak lip motion.

\subsubsection{Flow Matching with Speaker Switching}
The flow matching module is built upon the CosyVoice 3~\cite{CosyVoice3} backbone to reconstruct Mel-spectrograms from speech tokens.
However, the original flow matching mechanism conditions only on a single global speaker embedding, 
which limits its ability to perform speaker switching in dialogue and multi-speaker scenes.
To address speaker timbre switching in dialogue and multi-speaker settings, 
we propose a speaker switching concatenation strategy conditioned on $M$ reference speakers $ref^{M}$.
The universal speaker encoder~\cite{3D-Speaker} based on CAM++ is employed to extract audio speaker embeddings $E_{Spk}^M$ for $ref^{M}$.
Based on the timestamp-speaker tuples $\mathcal{T}^N$, the last silent speech token within the temporal neighborhood of each timestamp segment is identified via bidirectional search.
The corresponding speaker embedding $E_{Spk}\in E_{Spk}^M$ is then inserted immediately after the last silent token and before the onset of speech tokens, 
which are encoded into $E_{ST^*}^T$ using a causal upsampling encoder.
This design enables explicit speaker switching aligned with timestamp boundaries, as shown in Fig.~\ref{fig:FunCineForge}.

The resulting concatenated feature sequence is fed into a Diffusion Transformer (DiT)~\cite{DiT} as conditional context $\mathcal{C}_{flow}$ for flow matching training,
\begin{equation}
	\mathcal{L}_{Flow}(\theta) = \mathbb{E}_{u,Y_1}||\mathrm{DiT}_\theta \bigl(Y_u,u \mid \mathcal{C}_{flow}\bigr) - (Y_1-Y_0)||_2^2,
\end{equation}
where $\mathcal{C}_{flow}=\mathrm{Concat}\bigl(E_{ST^*}^T, E_{Spk}^M(\mathcal{T}^N)\bigr)$, $u\thicksim U(0,1)$, 
$Y_1\thicksim \mathcal{N}(\mathbf{0},\mathbf{I})$, 
$Y_u = (1-u)Y_0 + uY_1$.
During training, the conditioning $\mathcal{C}_{flow}$ is randomly dropped with a fixed probability to enable classifier-free guidance at inference stage.

\section{Experiments}
\subsection{Implementation Details}
For data preprocessing, all videos are standardized to the MP4 format using libx264 and libmp3lame encoders,
and the opening and ending credits of television episodes are removed.
Video frames are sampled at 25 fps, and all audio is resampled to 16 kHz.
The speech tokenizer uses a FSQ codebook of size 6,560, and the frame-index codebook size is 1,500.
The MLLM and the flow matching module of the dubbing model are trained separately on the CineDub-CN dataset for more than 20 epochs.
They are initialized from the CosyVoice3-0.5B checkpoint, which already demonstrates strong speech naturalness.
Training is conducted on 8 NVIDIA A100 GPUs with 80-core CPUs,
using the AdamW optimizer with a peak learning rate of 1e-4, 2,000 warm-up steps, 
and a batch size of 2e+4 tokens.

For evaluation, we select several open-source SOTA baselines specifically designed for movie dubbing,
and use their official checkpoints trained on the V2C-Animation, Chem, or GRID datasets for inference.
For instruction-driven methods, DeepDubber-V1 and InstructDubber,
we modify their implementations and retrain them on the CineDub-CN dataset to enable comparison under the same data setting.
In addition, we train the FunCineForge dubbing model on the combined V2C-Animation, Chem, and GRID datasets 
using only monologue scene with empty clue and timestamp-speaker annotations, 
in order to analyze dataset quality and to compare with existing baselines.

\begin{table*}
    \centering
    \resizebox{\linewidth}{!}{
    \begin{tabular}{l|rrrrrrrrrrr}
        \toprule
        \multicolumn{12}{c}{\textbf{FunCineForge Dubbing Model}} \\
        \midrule
        Scene         & MCD-DTW$\downarrow$ & MCD-DTW-SL$\downarrow$  & CER(\%)$\downarrow$ & WER(\%)$\downarrow$   & UTMOS$\uparrow$   & LSE-C$\uparrow$   & LSE-D$\downarrow$ & SPK-TL$\downarrow$    & SPK-SIM(\%)$\uparrow$ & EMO-SIM(\%)$\uparrow$ & ES-MOS$\uparrow$\\
        \midrule
        \multicolumn{12}{l}{\textbf{CineDub-CN Dataset (Corrected)}} \\
        \midrule
        GT            & 0.00              & 0.00                      & \textbf{0.94}       & -                     & 3.86              & 8.35              & 5.95              & 0.000                 & 100.00                & 100.00                & 3.94  \\
        Monologue     & \textbf{4.52}     & \textbf{4.58}             & 1.55                & -                     & \textbf{3.98}     & \textbf{8.72}     & \textbf{3.82}     & 0.088                 & \textbf{76.50}        & \textbf{74.50}        & 3.80  \\
        Narration     & 4.65              & 4.78                      & 2.23                & -                     & 3.92              & -                 & -                 & \textbf{0.062}        & 74.51                 & 59.92                 & 3.64  \\
        Dialogue      & 4.91              & 5.06                      & 3.14                & -                     & 3.85              & 8.55              & 5.07              & 0.090                 & 68.05                 & 70.47                 & 3.86  \\
        Multi-speaker & 5.21              & 5.48                      & 3.37                & -                     & 3.80              & 8.04              & 6.65              & 0.073                 & 67.75                 & 62.69                 & \textbf{4.03} \\
        \midrule
        \multicolumn{12}{l}{\textbf{CineDub-CN Dataset (Uncorrected)}} \\
        \midrule
        GT            & 0.00              & 0.00                      & 4.53                & -                     & 3.85              & 7.98              & 8.18              & 0.000                 & 100.00                & 100.00                & 3.73  \\
        Monologue     & 5.88              & 5.92                      & 3.80                & -                     & 3.89              & 8.30              & 6.95              & 0.122                 & 71.30                 & 72.57                 & 3.75  \\
        Narration     & 5.81              & 6.05                      & 5.56                & -                     & 3.82              & -                 & -                 & 0.104                 & 69.44                 & 62.12                 & 3.70  \\
        Dialogue      & 6.06              & 6.53                      & 10.67               & -                     & 3.71              & 7.82              & 9.20              & 0.169                 & 66.43                 & 65.25                 & 3.55  \\
        Multi-speaker & 6.41              & 6.58                      & 12.62               & -                     & 3.65              & 7.24              & 10.88             & 0.145                 & 65.80                 & 67.30                 & 3.89  \\
        \midrule
        \multicolumn{12}{l}{\textbf{V2C-Animation + Chem + GRID Dataset}} \\
        \midrule
        GT            & 0.00              & 0.00                      & -                   & 24.23                 & 3.70              & 7.05              & 7.55              & 0.000                 & 100.00                & 100.00                & -  \\
        Monologue     & 6.58              & 6.80                      & -                   & \textbf{8.46}         & 3.76              & 7.50              & 9.96              & 0.223                 & 68.80                 & 64.25                 & -  \\
        \bottomrule
        \toprule
        \multicolumn{12}{c}{\textbf{Baselines (Monologue Scene)}} \\
        \midrule
        Methods                     & MCD-DTW$\downarrow$   & MCD-DTW-SL$\downarrow$    & CER(\%)$\downarrow$   & WER(\%)$\downarrow$   & UTMOS$\uparrow$   & LSE-C$\uparrow$   & LSE-D$\downarrow$ & SPK-TL$\downarrow$    & SPK-SIM(\%)$\uparrow$ & EMO-SIM(\%)$\uparrow$ & ES-MOS$\uparrow$\\
        \midrule
        \multicolumn{12}{l}{\textbf{CineDub-CN Dataset (Corrected)}} \\
        \midrule
        DeepDubber-V1~\cite{DeepDubber-V1}  & 5.25          & 5.33                      & 6.05                  & -                     & 3.70              & 7.86              & 9.24              & 0.140                 & 71.61                 & 65.57                 & 3.70  \\
        InstructDubber~\cite{InstructDubber}& 5.05          & 5.23                      & 3.84                  & -                     & 3.82              & 8.08              & 7.93              & 0.156                 & 74.53                 & 72.86                 & 3.83  \\
        \midrule
        \multicolumn{12}{l}{\textbf{V2C-Animation + Chem + GRID Dataset}} \\
        \midrule
        Speaker2Dubber~\cite{Speaker2Dubber}& 9.80          & 10.03                     & -                     & 32.12                 & 3.42              & 5.63              & 12.58             & 0.307                 & 63.05                 & 46.33                 & -  \\
        StyleDubber~\cite{StyleDubber}      & 8.52          & 8.63                      & -                     & 29.45                 & 3.70              & 6.03              & 12.05             & 0.213                 & 64.80                 & 49.52                 & -  \\
        ProDubber~\cite{ProDubber}          & 6.60          & 6.72                      & -                     & 11.02                 & 3.82              & 5.89              & 10.45             & 0.149                 & 70.50                 & 58.19                 & -  \\
        EmoDubber~\cite{EmoDubber}          & 8.83          & 8.89                      & -                     & 22.14                 & 3.84              & 7.80              & 7.67              & 0.160                 & 65.14                 & 68.78                 & -  \\
        DeepDubber-V1~\cite{DeepDubber-V1}  & 7.46          & 7.55                      & -                     & 24.28                 & 3.68              & 6.33              & 10.07             & 0.208                 & 67.17                 & 60.35                 & 3.32  \\
        InstructDubber~\cite{InstructDubber}& 6.69          & 6.89                      & -                     & 14.20                 & 3.70              & 6.50              & 9.68              & 0.166                 & 69.55                 & 70.02                 & 3.70  \\
        \bottomrule
    \end{tabular}
    }
\caption{The upper part reports evaluation results of the FunCineForge dubbing model trained on the CineDub-CN dataset (with and without Multimodal CoT Correction) and on the combined V2C-Animation, Chem, and GRID datasets.
The synthesized speech in diverse scenes is evaluated against ground-truth on the corresponding test sets of each dataset.
GT denotes evaluating the intrinsic quality of each dataset in terms of text quality, speech quality, audio-visual alignment, and audio-clue emotional and style consistency.
The lower part reports baseline results under the monologue scene for comparison.
}
\label{tab:correction}
\end{table*}

\subsection{Evaluation Metrics}
We adopt state-of-the-art objective and subjective evaluation metrics to 
comprehensively assess dataset quality, dubbing quality, audio-visual alignment, and other relevant aspects.

\paragraph{MCD-DTW \& MCD-DTW-SL.}
Mel Cepstral Distortion with Dynamic Time Warping (MCD-DTW) and its variant MCD-DTW-SL~\cite{V2C}, 
which incorporates duration-aware weighting, 
measure the difference between synthesized speech and ground-truth recordings.

\paragraph{UTMOS.}
UTMOS~\cite{UTMOS} is a 5-point speech mean opinion score (MOS) predictor to assess the naturalness of synthesized speech.

\paragraph{LSE-D \& LSE-C.}
To accurately measure lip sync, 
we use Lip Sync Error Distance and Lip Sync Error Confidence~\cite{LSE-D}.
Both metrics are computed based on the pre-trained SyncNet,
which is widely adopted in lip-reading task.

\paragraph{SPK-TL.}
Given the timestamp-speaker tuples $\mathcal{T}^N=\{(t_i^{\text{start}}, s_i, g_i, a_i, t_i^{\text{end}})\}_{i=1}^N$ from ground-truth videos,
the synthesized speech is expected to follow the provided intervals $\mathcal{I}_i=[t_i^{\text{start}}, t_i^{\text{end}})$ during speaker switching.
We extract speech-active regions from the synthesized speech using VAD 
and define SPK-TL to quantify speaker truncation and leakage
by measuring the temporal mismatch between $\hat{\mathcal{I}_i}$ of synthesized speech and $\mathcal{I}_i$,
\begin{equation}
    \mathrm{SPK-TL}=\frac{1}{2}-\frac{1}{2N}\sum_{i=1}^N\left(\frac{|\hat{\mathcal{I}_i}\cap\mathcal{I}_i|}{|\mathcal{I}_i|}-\frac{|\hat{\mathcal{I}_i}\setminus\mathcal{I}_i|}{|\hat{\mathcal{I}_i}|}\right),
\end{equation}
where lower values indicate more accurate alignment and speaker switching,
while values closer to 1 correspond to severe cross-speaker leakage.

\paragraph{SPK-SIM.}
Speaker cosine similarity is used to evaluate speaker timbre similarity between the synthesized speech 
within each speaker's active time interval and the corresponding reference speech.

\paragraph{EMO-SIM \& ES-MOS.}
Emotional learning depends on the input face images and clue instructions.
We employ emotion2vec\footnote{https://github.com/ddlBoJack/emotion2vec} to 
extract emotion vectors from synthesized speech and ground-truth recordings,
and compute the cosine similarity (EMO-SIM) between them.
In addition, we conduct a subjective 5-point MOS on emotional and style consistency (ES-MOS),
where human raters evaluate how well the speech matches the clue description content.

\paragraph{CER \& WER.}
Character Error Rate and Word Error Rate are used to measure pronunciation accuracy, 
using Whisper-large-v3\footnote{https://huggingface.co/openai/whisper-large-v3} as the ASR model.

\subsection{CineDub-CN Dataset and Quality Evaluation}
To train the dubbing model and evaluate the quality of the dubbing dataset produced by the dataset pipeline,
over 200 raw Chinese television series totaling more than 6,000~hours and covering diverse television genres are carefully collected,
with strict selection criteria including non documentaries, standard pronunciation, prominent vocal tracks, reduced colloquial speech,
and unobstructed faces (i.e., without masks or veils).
Using the pipeline, 1,559,172 samples are constructed, totaling 7.2~TB, with an average clip length of 11.02~seconds
and over 4,700~hours of effective speech.
Non-neutral emotional speech accounts for 41.8\% of the dataset, and
the average clue length reaches 62 words with a variance of 25, covering character attributes, speaking styles, and emotional variations,
indicating strong diversity and robust multi-scale clue annotations.
Additional dataset statistics are provided in Fig.~\ref{fig:statistics}.
We select four samples from each television series, corresponding to the four scene categories, to construct the test set.
Part of CineDub-CN dataset is released at here\footnote{https://anonymous.4open.science/w/FunCineForge}.

Table~\ref{tab:correction} (upper part) compares the intrinsic quality reported in the GT rows 
for the CineDub-CN dataset (with and without Multimodal CoT Correction) and the combined V2C-Animation, Chem, and GRID datasets,
as well as the performance of the FunCineForge dubbing model trained on these datasets.
Results show that models trained on these three small-scale, limited-annotation datasets 
tend to underfit lip sync, leading to lower alignment accuracy, and produce speech with lower naturalness.
In contrast, training on our CineDub-CN dataset
significantly improves pronunciation quality and lip sync.
Moreover, training on the corrected CineDub-CN dataset leads to clear improvements across all metrics 
compared with the uncorrected version, especially CER, LSE-D, and SPK-TL,
demonstrating that the proposed bidirectional verification mechanism 
between lightweight specialized models and general-purpose MLLM 
significantly reduces word and punctuation errors in the training data,
and alleviates duplicated and missing speaker identities in speaker diarization predictions, 
thereby substantially improving dataset quality.

\subsection{Comparison with SOTA and Scene Analysis}
As shown in Table~\ref{tab:correction}, to verify the effectiveness of the proposed FunCineForge dubbing model, 
we conduct comparisons under a unified monologue setting, where all SOTA methods are evaluated on the same test samples.
On the three English dubbing datasets, 
our model achieves consistently superior performance in terms of speech quality, WER and alignment accuracy.
In addition, we compare our model with two MLLM-based dubbing models trained on the CineDub-CN dataset
and evaluated on the same monologue test set.
The results demonstrate that our model achieves significant improvements across all evaluation metrics, 
highlighting the effectiveness of the proposed model architecture and design.

Furthermore, we evaluate the performance of FunCineForge dubbing model across four scene categories.
The results indicate that our model achieves higher speech naturalness and alignment accuracy in monologue and narration scenes.
In dialogue and multi-speaker scenes, which are inherently more challenging, 
the model maintains strong temporal alignment, accurate lip sync, and effective timbre switching.
With longer facial sequences and richer clue instructions, 
it further exhibits more accurate emotional expression and improved instruction following capability.

\begin{table}
    \centering
    \resizebox{\linewidth}{!}{
    \begin{tabular}{l|l|rrrrrrrrr}
        \toprule
        Methods                             & Scene         & LSE-C$\uparrow$   & LSE-D$\downarrow$ & SPK-TL$\downarrow$    & SPK-SIM(\%)$\uparrow$ & ES-MOS$\uparrow$  \\
        \midrule
        \multirow{4}{*}{FunCineForge}       & Monologue     & \textbf{8.72}     & \textbf{3.82}     & 0.088                 & \textbf{76.50}        & 3.80 \\
                                            & Narration     & -                 & -                 & \textbf{0.062}        & 74.51                 & 3.64 \\
                                            & Dialogue      & 8.55              & 5.07              & 0.090                 & 68.05                 & 3.86 \\
                                            & Multi-Speaker & 8.04              & 6.65              & 0.073                 & 67.75                 & \textbf{4.03} \\
        \midrule                                        
        \multirow{4}{*}{w/o $\mathcal{T}^N$}& Monologue     & 8.05              & 7.53              & 0.150                 & 72.20                 & 3.70 \\
                                            & Narration     & -                 & -                 & 0.164                 & 70.85                 & 3.60 \\
                                            & Dialogue      & 5.98              & 12.94             & 0.313                 & 62.11                 & 3.74 \\
                                            & Multi-Speaker & 5.58              & 13.12             & 0.340                 & 57.40                 & 3.86 \\
        \midrule                                        
        \multirow{4}{*}{w/o $\mathcal{L}_{Lip}$}& Monologue & 8.55              & 4.01              & 0.092                 & 76.02                 & 3.75 \\
                                            & Narration     & -                 & -                 & 0.063                 & 74.43                 & 3.67 \\
                                            & Dialogue      & 8.41              & 5.88              & 0.095                 & 68.17                 & 3.84 \\
                                            & Multi-Speaker & 7.80              & 7.45              & 0.078                 & 67.88                 & 3.96 \\
        \midrule                                        
        \multirow{4}{*}{w/o SSC}            & Monologue     & 8.68              & 3.89              & 0.090                 & 73.40                 & 3.73 \\
                                            & Narration     & -                 & -                 & 0.062                 & 71.95                 & 3.55 \\
                                            & Dialogue      & 8.50              & 5.32              & 0.094                 & 58.85                 & 3.63 \\
                                            & Multi-Speaker & 8.01              & 6.70              & 0.075                 & 52.30                 & 3.81 \\

        \bottomrule
    \end{tabular}
    }
\caption{Results of ablation study on the CineDub-CN dataset.}
\label{tab:2}
\end{table}

\subsection{Ablation Studies}
To analyze the contribution of the key components in the our dubbing model,
we conduct ablation studies on the CineDub-CN dataset.
Results are summarized in Table~\ref{tab:2}.

\paragraph{Effectiveness of timestamp-speaker tuples $\mathcal{T}^N$.}
Removing the timestamp-speaker tokenizer, the tuple inputs $\mathcal{T}^N$ and loss $\mathcal{L}_{VA}$,
causes clear degradation in temporal alignment.
Both SPK-TL and LSE-D increase substantially, especially in dialogue and multi-speaker scenes, where speaker truncation and leakage frequently occur.
This shows that facial features alone are insufficient for long sequences with frequent shot changes and speaker switching, 
and explicit temporal speaker supervision is essential for complex cinematic scenes.

\paragraph{Effectiveness of lip contrastive loss $\mathcal{L}_{Lip}$.}
When removing $\mathcal{L}_{Lip}$ while keeping facial embeddings, LSE-C decreases and LSE-D increases, indicating weaker lip sync.
Although facial features provide coarse lip motion cues, the lack of frame-level contrastive supervision leads to reduced fine-grained alignment accuracy.

\paragraph{Effectiveness of speaker switching concatenation (SSC).}
We remove the SSC strategy and condition the flow matching on a global reference speaker embedding as in CosyVoice3.
Speaker similarity drops slightly in monologue and narration scenes,
but degrades significantly in dialogue and multi-speaker scenes,
where the synthesized speech converges toward an averaged timbre.
Nevertheless, subjective evaluation shows that speaker attributes such as gender and age remain distinguishable,
likely due to the multi-task supervision in the CosyVoice3 speech tokenizer, which includes speaker age and gender prediction.

\section{Conclusion}
In this work, we propose an end-to-end dataset pipeline and an MLLM-based dubbing model designed for live-action cinematic dubbing.
We construct the first Chinese television dubbing dataset with rich annotations.
Multimodal CoT Correction improves dataset quality, while robust clue instructions enhance emotional consistency.
The frame-index codebook and dedicated MLLM supervision enable accurate alignment in complex scenes,
and the improved flow-matching design supports flexible speaker switching.
In future work, we will expand the dataset to multilingual settings and explore joint video-to-audio and speech generation.

\appendix

\section*{Ethical Statement}
All third-party tools and models used in this work are sourced from open-source projects and comply with their licenses. 
This work is conducted solely for research purposes, and released data samples are screened according to regulations. 
The demos are provided only for demonstration purposes and contain no sensitive or infringing information. 
All human participants involved in subjective evaluations are compensated under employment agreements. 
The experiments and datasets pose no ethical issues.


\bibliographystyle{named}
\bibliography{ijcai26}

\end{document}